%% file: main.tex
\documentclass[lettersize,journal]{IEEEtran} 
\usepackage{amsmath,amsfonts}
\usepackage{algorithmic}
\usepackage{algorithm}
\usepackage{array}
\usepackage{arydshln}
\usepackage[caption=false,font=normalsize,labelfont=sf,textfont=sf]{subfig}

\usepackage{textcomp}
\usepackage{stfloats}
\usepackage{url}
\usepackage{verbatim}
\usepackage{graphicx}
\usepackage{cite}

\usepackage{multirow}
\usepackage{colortbl}
\usepackage{booktabs}
\usepackage{arydshln}

\newcolumntype{C}[1]{>{\centering\arraybackslash}m{#1}}

\usepackage[table]{xcolor}
\definecolor{rankone}{RGB}{226,239,218}     
\definecolor{ranktwo}{RGB}{221,235,247}     
\definecolor{rankthree}{RGB}{252,228,214}   

\definecolor{psnrone}{RGB}{198,224,180}     
\definecolor{psnrtwo}{RGB}{189,215,238}     
\definecolor{psnrthree}{RGB}{248,203,173}   

\definecolor{ssimone}{RGB}{235,244,228}     
\definecolor{ssimtwo}{RGB}{234,242,250}     
\definecolor{ssimthree}{RGB}{253,237,227}   

\definecolor{lpipsone}{RGB}{232,222,248}    
\definecolor{lpipstwo}{RGB}{220,235,247}    
\definecolor{lpipsthree}{RGB}{245,228,215}  

\begin{document}

\title{RMGS-SLAM: Real-time Multi-sensor Gaussian Splatting SLAM}

\author{Dongen Li\textsuperscript{1,*}, Yi Liu\textsuperscript{1,*}, Junqi Liu\textsuperscript{3,*}, Zewen Sun\textsuperscript{1,*}, Zefan Huang\textsuperscript{1,*}, Shuo Sun\textsuperscript{4}, Jiahui Liu\textsuperscript{2}, Chengran Yuan\textsuperscript{1}, Hongliang Guo\textsuperscript{3}, Francis E.H. Tay\textsuperscript{1}, and Marcelo H. Ang Jr.\textsuperscript{1}%

\thanks{\textsuperscript{*}Dongen Li, Yi Liu, Junqi Liu, Zewen Sun, and Zefan Huang contributed equally to this work.}%
\thanks{\textsuperscript{1}Dongen Li, Yi Liu, Zewen Sun, Zefan Huang, Chengran Yuan, Francis E.H. Tay, and Marcelo H. Ang Jr. are with the Advanced Robotics Centre, National University of Singapore.}%
\thanks{\textsuperscript{2}Jiahui Liu is with the School of Aeronautics and Astronautics, Shanghai Jiao Tong University.}%
\thanks{\textsuperscript{3}Junqi Liu and Hongliang Guo are with the College of Computer Science, Sichuan University.}%
\thanks{\textsuperscript{4}Shuo Sun is with Singapore-MIT Alliance for Research and Technology.}}




\maketitle

\begin{abstract}
Achieving real-time Simultaneous Localization and Mapping (SLAM) based on 3D Gaussian splatting (3DGS) in large-scale real-world environments remains challenging, as existing methods still struggle to jointly achieve low-latency pose estimation, continuous 3D Gaussian reconstruction, and long-term global consistency. In this paper, we present a tightly coupled LiDAR-Inertial-Visual 3DGS-based SLAM framework for real-time pose estimation and photorealistic mapping in large-scale real-world scenes. The system executes state estimation and 3D Gaussian primitive initialization in parallel with global Gaussian optimization, enabling continuous dense mapping. To improve Gaussian initialization quality and accelerate optimization convergence, we introduce a cascaded strategy that combines feed-forward predictions with geometric priors derived from voxel-based principal component analysis. To enhance global consistency, we perform loop closure directly on the optimized global Gaussian map by estimating loop constraints through Gaussian-based Generalized Iterative Closest Point registration, followed by pose-graph optimization. We also collect challenging large-scale looped outdoor sequences with hardware-synchronized LiDAR-camera-IMU and ground-truth trajectories for realistic evaluation. Extensive experiments on both public datasets and our dataset demonstrate that the proposed method achieves a state of the art among real-time efficiency, localization accuracy, and rendering quality across diverse real-world scenes.
\end{abstract}


\section{Introduction}

\IEEEPARstart{S}{imultaneous} Localization and Mapping (SLAM) enables robots to estimate their pose while reconstructing the surrounding environment~\cite{cadena2017past}. Conventional SLAM systems typically build geometry-centric maps~\cite{zhang2014loam,mur2017orb,zheng2024fast}, which offer limited support for photorealistic scene representation needed by immersive visual applications and downstream robotic tasks such as navigation and grasping. Meanwhile, 3D Gaussian splatting (3DGS)~\cite{kerbl3Dgaussians} represents a scene using explicit anisotropic Gaussian primitives and enables efficient differentiable rasterization, providing an appearance-aware, continuous, and directly renderable map representation. These properties make 3DGS particularly attractive for SLAM, as it provides both a geometric reference for pose estimation and a photorealistic representation for interaction and perception.


However, high-quality 3DGS reconstruction~\cite{kerbl3Dgaussians, zhou2024drivinggaussian, huang20242d} has mainly been explored in offline settings, where full observations are available and computation is less constrained. Extending 3DGS to real-time SLAM is substantially more challenging, since high-quality 3DGS reconstruction typically depends on iterative optimization, whereas real-time SLAM systems require both pose estimation and mapping to keep pace with incoming sensor streams. Although recent 3DGS-based SLAM systems have made encouraging progress, many existing methods~\cite{yan2023gs_gsslam, matsuki2024gaussian_monoGs, keetha2024splatam, deng2025vpgs} remain closer to online SLAM than to real-time SLAM. While supporting incremental pose estimation and Gaussian map updates, they generally struggle to sustain both processes under high-frequency continuous sensor inputs. By contrast, several recent LiDAR-assisted methods~\cite{hong2025gs, phan2025fusiongs,lang2025gaussian_gaussian_lic,lang2025gaussian2_gaussian_lic2,xie2025gs} better keep pace with continuous inputs, yet Gaussian initialization in these systems still often relies mainly on LiDAR observations with limited geometric assumptions, requiring further online refinement. In addition, explicit loop closure is often omitted to preserve runtime efficiency, which limits global consistency in large-scale environments. This paper aims to address these limitations with the following contributions:

\begin{itemize}
    \item We propose a real-time, tightly coupled LiDAR-Inertial-Visual (LIV) 3DGS-SLAM framework with explicit loop closure. The system supports non-blocking dense mapping by jointly executing pose estimation and Gaussian initialization in parallel with global Gaussian map optimization.

    \item We introduce a cascaded Gaussian initialization strategy that combines feed-forward predictions with geometric priors. It improves Gaussian primitive initialization quality, accelerates optimization convergence, and facilitates real-time Gaussian mapping.
    
    \item We develop a 3DGS-based loop-closure method for large-scale scenes. It extracts target Gaussian sets from the global Gaussian map and constructs loop constraints for global optimization to suppress long-term drift.
    
    \item We present a new real-world benchmark with synchronized solid-state LiDAR, camera, and IMU streams, together with ground truth for outdoor road scenes. It enables comprehensive evaluation of 3DGS-based SLAM systems under practical conditions.

    \item We validate the effectiveness of RMGS-SLAM in real-world scenarios through both our own sequences and publicly available datasets.
\end{itemize}

\section{RELATED WORK}
\subsection{3DGS for Scene Reconstruction}
3DGS~\cite{kerbl3Dgaussians} enables efficient, high-fidelity novel-view synthesis by representing scenes as 3D Gaussians and rendering them with a differentiable tile-based rasterizer. Conventional 3DGS lacks sufficient geometric constraints in real-world scenes. Accordingly, recent methods~\cite{zhou2024drivinggaussian, huang20242d} incorporate temporal tracking and structural constraints to reduce volumetric overlap and improve surface modeling. These optimization-based methods still require prolonged per-scene training.

To bypass prolonged per-scene optimization, generalizable feed-forward models predict Gaussians from sparse images. Pose-required models use known camera poses to fuse multi-view features and regress depth distributions to anchor Gaussians~\cite{xu2025depthsplat}. To avoid brittle offline pose estimation under sparse views, pose-free models jointly infer scene geometry and camera registration, reconstructing geometrically consistent scenes from unposed images~\cite{g3splat}. These advances suggest that feed-forward Gaussian prediction provides an efficient and increasingly generalizable paradigm for 3DGS initialization.

\subsection{3DGS-based SLAM}

\input{Figures_tex/overview_figure}

Recent works have extended 3DGS to SLAM by exploiting its explicit and differentiable representation for joint pose estimation and dense reconstruction. Early 3DGS-based SLAM systems were primarily developed for visual SLAM, including GS-SLAM~\cite{yan2023gs_gsslam}, which combines adaptive densification with coarse-to-fine tracking, MonoGS~\cite{matsuki2024gaussian_monoGs} for monocular RGB-only SLAM, and SplaTAM~\cite{keetha2024splatam} for efficient RGB-D mapping. 
Nevertheless, visual 3DGS-based SLAM remains limited in unconstrained real-world environments, where viewpoint variation, illumination changes, and imperfect geometry often degrade both tracking robustness and reconstruction quality.

To improve geometric perception and pose robustness, recent works incorporate LiDAR into 3DGS-based SLAM. Loosely coupled methods~\cite{deng2025vpgs,cheng2025tls,hong2024liv,xiao2024liv} typically rely on a dominant sensing modality or perform asynchronous multi-sensor pose estimation while reconstructing the Gaussian map. Although such designs simplify system integration, the loose or asynchronous interaction between tracking and mapping makes it difficult to maintain real-time performance under high-rate sensor inputs.

By contrast, tightly coupled methods jointly fuse multi-modal measurements within a unified estimator and integrate them with Gaussian mapping to improve spatiotemporal consistency. Gaussian-LIC~\cite{lang2025gaussian_gaussian_lic,lang2025gaussian2_gaussian_lic2}, LVI-GS~\cite{zhao2025lvi}, and FusionGS-SLAM~\cite{phan2025fusiongs} demonstrate the potential of deeper multi-modal integration, yet maintaining both real-time performance and global consistency remains challenging in large-scale environments.

Several systems are explicitly designed to keep 3DGS reconstruction in step with incoming sensor streams by restricting the representation size or optimization scope. GS-LIVM~\cite{xie2025gs} employs Voxel-GPR to regularize sparse and uneven LiDAR observations for more geometry-aware Gaussian primitives initialization, but still relies on continued optimization after the input stream ends to achieve satisfactory reconstruction quality. GS-LIVO~\cite{hong2025gs} tightly couples Gaussian mapping with a sequential-update iterative error state Kalman filter (IESKF), and adopts sliding-window to optimize local Gaussians, reducing optimization pressure, but limited real-time iteration budgets restrict the recovery of fine photorealistic details.

Overall, existing frameworks still struggle to simultaneously keep pace with incoming sensor streams, achieve robust loop closure in large-scale scenes, and maintain high-quality Gaussian map optimization. These limitations motivate a framework that combines real-time Gaussian mapping with Gaussian-based loop closure for large-scale consistency.

\section{METHODOLOGY}

The proposed system follows a four-module design, as illustrated in Fig.~\ref{fig:system_overview}. We first present the geometric fitting process based on voxel-based principal component analysis (voxel-PCA) in Sec.~\ref{sec:voxel}, followed by the cascaded 3D Gaussian primitive initialization in Sec.~\ref{sec:init}, the asynchronous Gaussian optimization in Sec.~\ref{sec:opt}, and the Gaussian-based loop closure in Sec.~\ref{sec:loop}.

\subsection{Voxel-PCA Geometric Fitting}
\label{sec:voxel}

To provide structured geometric priors for Gaussian initialization, we employ a voxel-PCA geometric fitting scheme. We partition the latest front-end processed colored point cloud in the world frame into voxels using a hash-indexed octree structure~\cite{yuan2022efficient, zheng2024fast}. For each voxel, we compute the mean $\boldsymbol{\mu}$ and covariance $\mathbf{C}$, and perform eigen-decomposition to obtain ordered eigenvalues $\lambda_{\min} \leq \lambda_{\mathrm{mid}} \leq \lambda_{\max}$ with corresponding eigenvectors $\mathbf{v}_{\min}$, $\mathbf{v}_{\mathrm{mid}}$, and $\mathbf{v}_{\max}$, forming an orthonormal eigenvector matrix $\mathbf{V}$.


Geometry is classified by the eigenvalue distribution to capture both planar and line-like local structures, avoiding sole reliance on planar fitting that may oversimplify or miss slender linear structures. Here, $\tau_p$ and $\tau_l$ denote the thresholds for planar and linear classification, and are fixed across all experiments. Specifically, if $\lambda_{\min} < \tau_p$, the structure is classified as planar with normal $\mathbf{v}_{\min}$; if $\lambda_{\max}/\lambda_{\min} > \tau_l$, it is classified as linear with direction $\mathbf{v}_{\max}$; otherwise, it is treated as unreliable.

To jointly characterize local geometry and its uncertainty, we define a geometric descriptor $\mathbf{g} = [\mathbf{v}_k^{\top},\; \boldsymbol{\mu}^{\top}]^{\top} \in \mathbb{R}^6$, where $\mathbf{v}_k \in \{\mathbf{v}_{\min}, \mathbf{v}_{\max}\}$, with $\mathbf{v}_k = \mathbf{v}_{\min}$ for planar structures and $\mathbf{v}_k = \mathbf{v}_{\max}$ for linear structures.

The uncertainty of the geometric descriptor is propagated from point-level covariances within each voxel. Each point $\mathbf{p}_i^{W}$ is associated with a world-frame covariance $\boldsymbol{\Sigma}_i^{W} \in \mathbb{R}^{3 \times 3}$ that models measurement noise and pose uncertainty. Following~\cite{yuan2022efficient, 
liu2021balm}, the Jacobian $\mathbf{J}_i$ is written as
\begin{equation}
\begin{aligned}
\mathbf{J}_i
&=
\begin{bmatrix}
\mathbf{V}\mathbf{F}_i \\
\dfrac{1}{N}\mathbf{I}_3
\end{bmatrix},
\qquad
\mathbf{F}_i
=
\begin{bmatrix}
\mathbf{f}_i^{(\mathrm{min})} \\
\mathbf{f}_i^{(\mathrm{mid})} \\
\mathbf{f}_i^{(\mathrm{max})}
\end{bmatrix}, \\
\mathbf{f}_i^{(m)}
&=
\begin{cases}
\dfrac{(\mathbf{p}_i^W - \boldsymbol{\mu})^\top}{N(\lambda_k - \lambda_m)}
\left(
\mathbf{v}_m \mathbf{v}_k^\top + \mathbf{v}_k \mathbf{v}_m^\top
\right), & m \neq k, \\[6pt]
\mathbf{0}_{1\times 3}, & m = k,
\end{cases}
\end{aligned}
\end{equation}
where $N$ denotes the number of points contained in the current voxel, and $m \in \{\mathrm{min}, \mathrm{mid}, \mathrm{max}\}$. The descriptor covariance is then computed as
\begin{equation}
\boldsymbol{\Sigma}_{\mathbf{g}}
=
\sum_{i=1}^{N}
\mathbf{J}_i \boldsymbol{\Sigma}_i^W \mathbf{J}_i^{\top}.
\end{equation}




Geometric reliability is quantified by the trace of the directional covariance, and descriptors with large uncertainty are marked as unreliable. Each point is then assigned voxel-level attributes $\mathbf{g}_i
=
\left(
\mathbf{p}_i^W,\,
\mathbf{R}_{\text{pca}},\,
\mathbf{s}_{\text{pca}},\,
\mathbf{c}_i,\,
m_i
\right)$, where $\mathbf{c}_i \in \mathbb{R}^3$ is the color and $m_i \in \{0,1\}$ denotes geometric reliability. For reliable structures, $\mathbf{R}_{\text{pca}}$ is given by the PCA eigenvectors and $\mathbf{s}_{\text{pca}}$ is computed from the logarithm of the square roots of the corresponding eigenvalues.

This results in a compact geometric representation with explicit orientation, scale, and reliability indicators, which serves as a structured prior for subsequent initialization.

\subsection{Cascaded 3D Gaussian Primitive Initialization}
\label{sec:init}

From the synchronized front-end output stream, frames are periodically selected as keyframes based on a frame-index gap. Each keyframe is associated with a corrected pose estimate, an RGB image, a projected depth image, and a voxel-PCA geometric prior. A subset of keyframes is further designated as loopframes when the relative motion with respect to the latest loopframe exceeds predefined translation or rotation thresholds. Successive loopframes partition the keyframe sequence into local segments. When a new loopframe is created, observations from the keyframes between the previous and current loopframes are aggregated to initialize a new batch of Gaussians for that segment.

Each new loopframe closes a local segment with the previous one. We aggregate the 3D points observed within this segment and instantiate one Gaussian per point, initializing its world-frame mean at the point location and its DC component from the point color. The remaining attributes, including scale, rotation, opacity, and higher-order Spherical Harmonics (SH) coefficients, are collectively denoted by $\Theta(\mathbf{p})$ and initialized through a cascaded strategy that combines feed-forward predictions with voxel-PCA geometric priors to improve initialization quality.

\begin{equation}
\Theta(\mathbf{p})
=
\begin{cases}
\Theta_{\text{model}}(\mathbf{p}), & \text{valid model prediction}, \\
\Theta_{\text{pca}}(\mathbf{p}), & \text{invalid prediction} \wedge \text{geo-reliable}, \\
\Theta_{\text{heur}}(\mathbf{p}), & \text{otherwise}.
\end{cases}
\end{equation}

\input{Figures_tex/cascaded_init}

To obtain the model-based attributes  $\Theta_{\text{model}}(\mathbf{p})$, the current and previous loopframe images are processed by a pre-trained feed-forward model (FFM) that predicts per-pixel Gaussian attribute maps. In our implementation, this branch is instantiated with G$^3$Splat~\cite{g3splat}, although the proposed framework is not tied to any specific FFM. For each instantiated 3D point, we project it into both views and assign to it the predicted attributes at the corresponding pixel location. If the point is visible in both views, the attributes from the closer view are selected (as shown in Fig.~\ref{fig:cascaded_init}~(i)). To handle occlusion and projection conflicts, only the closest point is retained at each pixel, and the attributes are obtained by bilinear interpolation from the dense prediction maps (as shown in Fig.~\ref{fig:cascaded_init}~(ii)). The predicted rotation, represented as a quaternion, is then transformed from the camera frame to the world frame using the current camera pose (as illustrated in Fig.~\ref{fig:cascaded_init}~(iii)). Since the network predicts anisotropic proportions rather than absolute scales, we normalize the predicted scale to retain only its axis-wise shape ratio, denoted by $\mathbf{a}$, and recover the output Gaussian scale in log-space, denoted by $\mathbf{s}_{\text{out}}$, by anchoring it to the LiDAR depth using the ratio between the point depth $d$ and the camera focal length $f$:
\begin{equation}
\mathbf{s}_{\text{out}} =
\log\!\left(
(d / f) \cdot \mathbf{a}
\right).
\end{equation}

Under the PCA-based branch $\Theta_{\text{pca}}$, the voxel-PCA prior provides rotation and anisotropic scale in log-space. The rotation is directly derived from the PCA eigenvectors. To avoid over-stretched Gaussians in uncertain regions, the PCA-derived scale is further adjusted by a bounded dynamic factor $\beta$, determined by the ratio of points in the current accumulated segment that are geometrically reliable but have invalid model-predicted parameters to the total
number of accumulated points. The final scale is computed as
\begin{equation}
\mathbf{s}_{\text{out}} = \mathbf{s}_{\text{pca}} + \log \beta,
\end{equation}
where $\log \beta$ is added to each component of $\mathbf{s}_{\text{pca}}$. 

Under the heuristic branch $\Theta_{\text{heur}}$, the rotation is set to identity, and the scale is initialized isotropically using the depth-to-focal-length ratio:

\begin{equation}
\mathbf{s}_{\text{out}} = \log\!\left(d / f\right).
\end{equation}

For the two non-model branches, $\Theta_{\text{pca}}$ and $\Theta_{\text{heur}}$, the opacity is initialized to a constant value for uniform contribution, and all higher-order SH coefficients are set to zero.

This cascaded design integrates feed-forward appearance priors with geometry-aware fallback mechanisms, enabling robust initialization under incomplete or unreliable predictions and producing high-quality 3D Gaussian primitives that facilitate faster convergence.

\subsection{Optimization}
\label{sec:opt}

Newly initialized Gaussians are inserted into the map, while redundant ones are filtered beforehand via a spatial proximity test in 3D Euclidean space to suppress duplicated structures in revisited regions.

Optimization is supervised by keyframe images. Over the committed keyframes, we maintain two temporal windows: a recent window associated with the latest Gaussian segments and a history window containing earlier views. At each iteration, views are randomly sampled from both windows to balance fast adaptation and long-range consistency. To preserve previously optimized regions, only Gaussians in the most recent $K$ segments are updated, while older segments remain frozen.

To ensure stable supervision, we use renderings of the current Gaussian map to identify reliably reconstructed regions. Specifically, a geometric support mask $\mathcal{M}_{\mathrm{geo}}$ is derived from accumulated opacity, and a conservative interior mask $\mathcal{M}_{\mathrm{int}}                   $ is obtained via morphological erosion:
\begin{equation}
\mathcal{M}_{\mathrm{int}}
=
\mathrm{Erode}(\mathcal{M}_{\mathrm{geo}}, r_{\mathrm{erode}}),
\end{equation}
where $\mathrm{Erode}(\cdot, r_{\mathrm{erode}})$ denotes binary erosion with radius $r_{\mathrm{erode}}$. All supervision terms are evaluated within $\mathcal{M}_{\mathrm{int}}$ to avoid unstable gradients in under-observed regions.

Color rendering follows standard differentiable Gaussian splatting via alpha compositing along each ray. Depth is computed using the same compositing weights:
\begin{equation}
D(\mathbf{x})
=
\sum_{i=1}^{N} z_i T_i \alpha_i(\mathbf{x}),
\end{equation}
where $N$ denotes the number of Gaussians considered in alpha compositing along the ray, $z_i$ denotes the camera-frame depth of the $i$-th Gaussian, and $T_i \alpha_i(\mathbf{x})$ defines its effective compositing weight along the ray.

The overall objective $\mathcal{L}$ is defined as
\begin{equation}
\mathcal{L}
=
\lambda_{\mathrm{rgb}} \mathcal{L}_{\mathrm{rgb}}
+
\lambda_{\mathrm{ssim}} \mathcal{L}_{\mathrm{ssim}}
+
\lambda_{\mathrm{depth}} \mathcal{L}_{\mathrm{depth}}.
\end{equation}
\begin{equation}
\mathcal{L}_{\mathrm{rgb}}
=
\frac{1}{|\mathcal{M}_{\mathrm{int}}|}
\sum_{\mathbf{x} \in \mathcal{M}_{\mathrm{int}}}
\left\|
\mathbf{C}(\mathbf{x}) - \tilde{\mathbf{C}}(\mathbf{x})
\right\|_1,
\end{equation}
\begin{equation}
\mathcal{L}_{\mathrm{ssim}}
=
1
-
\mathrm{SSIM}\!\left(
\mathbf{C}\!\mid_{\mathcal{M}_{\mathrm{int}}},
\tilde{\mathbf{C}}\!\mid_{\mathcal{M}_{\mathrm{int}}}
\right),
\end{equation}
\begin{equation}
\mathcal{L}_{\mathrm{depth}}
=
\frac{1}{|\mathcal{M}_{\mathrm{int}}|}
\sum_{\mathbf{x} \in \mathcal{M}_{\mathrm{int}}}
\left|
D(\mathbf{x}) - \tilde{D}(\mathbf{x})
\right|,
\end{equation}
where $\mathbf{C}(\mathbf{x})$ and $\tilde{\mathbf{C}}(\mathbf{x})$ denote the rendered and observed RGB values, and $D(\mathbf{x})$ and $\tilde{D}(\mathbf{x})$ denote the rendered and reference depths, respectively. The loss terms correspond to photometric, structural, and depth consistency.

During optimization, visible and unfrozen Gaussians are refined by a sparse gradient-based optimizer. Additionally, low-opacity Gaussians are pruned under a bounded removal ratio to suppress weak or redundant primitives.

\subsection{Gaussian-based Loop Closure}
\label{sec:loop}

To enforce global consistency in large-scale scenes, loop closure is performed directly on the global 3D Gaussian map. For each current loopframe, we search for historical loopframes based on spatial proximity and retain those sufficiently separated in time and index as loop candidate. Each candidate pair formed by the current loopframe and loop candidate defines a potential loop constraint, for which relative pose estimation is performed. Specifically, for a current loopframe $i$ and a historical loopframe $j$, registration is carried out between a source Gaussian set $\mathcal{G}_{i}^{\mathrm{src}}$ and a target Gaussian set $\mathcal{G}_{j}^{\mathrm{tar}}$.

The source set $\mathcal{G}_{i}^{\mathrm{src}}$ is formed by the Gaussians associated with the current loopframe, represented under its current corrected pose. To construct the target set $\mathcal{G}_{j}^{\mathrm{tar}}$, we extract Gaussians from the optimized global 3D Gaussian map that fall within the camera frusta of views around the historical loopframe $j$. This mitigates incompleteness and boundary inconsistency caused by optimization and pruning. 
To obtain a denser and more reliable target set, we use a set of neighboring keyframes around the historical loopframe $j$, denoted by $\mathcal{K}_j$, as query views. For each Gaussian $G_n$ in the global model with center $\boldsymbol{\mu}_n$, let $\mathbf{p}_{c,n}^{k}$ denote its coordinates in the camera frame of a surrounding keyframe $k \in \mathcal{K}_j$. A Gaussian is retained if there exists at least one keyframe $k \in \mathcal{K}_j$ such that its projection $\pi(\mathbf{p}_{c,n}^{k})$ lies within the image domain $\Omega^{k}$. To suppress distant Gaussians with weak contributions, we further discard Gaussians whose centers are farther than $d_{\max}$ from the world-frame position $\mathbf{t}_{w}^{k}$, and exclude Gaussians in the source set $\mathcal{G}_{i}^{\mathrm{src}}$. The final target set is defined as
\begin{equation}
\mathcal{G}_{j}^{\mathrm{tar}}
=
\left\{
\begin{aligned}
G_n \;\mid\;&
\exists\,k\in\mathcal{K}_j,\;
\pi\!\left(\mathbf{p}_{c,n}^{k}\right)\in\Omega^{k},\\
&
\|\boldsymbol{\mu}_n-\mathbf{t}_{w}^{k}\|_2<d_{\max},\;
G_n\notin\mathcal{G}_{i}^{\mathrm{src}}
\end{aligned}
\right\},
\end{equation}
where $G_n$ denotes the $n$-th Gaussian in the global model.

Given $\mathcal{G}_{i}^{\mathrm{src}}$ and $\mathcal{G}_{j}^{\mathrm{tar}}$, we estimate the loop constraint for pose-graph optimization via a Gaussian-based Generalized Iterative Closest Point (GICP) registration, which extends the GICP \cite{segal2009generalized} formulation to Gaussian primitives. To improve registration robustness, each Gaussian covariance is regularized into a planar covariance form. Specifically, for the original source and target covariances $\boldsymbol{\Sigma}_{m}^{\mathrm{src}}$ and $\boldsymbol{\Sigma}_{n}^{\mathrm{tar}}$, we replace their eigenvalue matrices with $\tilde{\boldsymbol{\Lambda}}=\mathrm{diag}(1,\,1,\,10^{-3})$ to obtain the regularized covariances $\tilde{\boldsymbol{\Sigma}}_{m}^{\mathrm{src}}$ and $\tilde{\boldsymbol{\Sigma}}_{n}^{\mathrm{tar}}$. Let $\mathbf{T}=(\mathbf{R},\mathbf{t})\in SE(3)$ denote the rigid transform that aligns the source Gaussian set $\mathcal{G}_{i}^{\mathrm{src}}$ to the target Gaussian set $\mathcal{G}_{j}^{\mathrm{tar}}$. For a correspondence $(m,n)$ between source and target Gaussians, the residual and its corresponding covariance are defined as
\begin{equation}
\mathbf{r}_{mn}(\mathbf{T})
=
\boldsymbol{\mu}_{n}^{\mathrm{tar}}
-
\mathbf{T}\boldsymbol{\mu}_{m}^{\mathrm{src}},
\end{equation}
\begin{equation}
\boldsymbol{\Sigma}_{mn}(\mathbf{T})
=
\tilde{\boldsymbol{\Sigma}}_{n}^{\mathrm{tar}}
+
\mathbf{R}\tilde{\boldsymbol{\Sigma}}_{m}^{\mathrm{src}}\mathbf{R}^{\top}.
\end{equation}
The loop relative pose between Gaussian sets is estimated by
\begin{equation}
\mathbf{T}_{ij}^{\mathrm{loop}}
=
\arg\min_{\mathbf{T} \in SE(3)}
\sum_{(m,n) \in \mathcal{C}}
\mathbf{r}_{mn}(\mathbf{T})^{\top}
\boldsymbol{\Sigma}_{mn}(\mathbf{T})^{-1}
\mathbf{r}_{mn}(\mathbf{T}),
\end{equation}
where $\mathcal{C}$ is the correspondence set. A loop candidate is accepted only if the optimization converges and the residual error falls below a predefined threshold.

The accepted alignment is subsequently converted into a loop edge and then jointly optimized with odometry constraints in a GTSAM-based~\cite{kaess2012isam2} pose graph. After pose-graph optimization, the updated corrected poses are propagated to the associated Gaussian segments in the global 3D Gaussian map, while the corresponding supervisory camera views are updated by the same rigid transform. This process enforces global consistency across the pose graph, the Gaussian map, and the associated views.

\section{EXPERIMENTS}
\subsection{Experimental Setup}

\textit{1) Datasets:} To evaluate the proposed method in realistic scenarios,  we first compare with representative sequences from the FAST-LIVO2 dataset~\cite{zheng2024fast} and the MARS-LVIG dataset~\cite{li2024mars}. Additionally, we collected our own outdoor driving datasets, named \textit{Driving1} and \textit{Driving2}, by building a strictly time-synchronized LiDAR-Camera-IMU device mounted on a golf buggy, comprising a Livox Avia LiDAR, a FLIR Blackfly S BFS-U3-51S5C camera, and an Arduino Mega microcontroller that provides a 10\,Hz PWM trigger for hardware-level synchronization. Ground truth is provided by an Xsens MTi-680G RTK GNSS/INS device, allowing the collected dataset to complement existing solid-state LIV benchmarks with large-scale looped outdoor road scenes. In total, we test our method on sequences with trajectory lengths ranging from 44\,m to 1,988\,m, enabling a thorough evaluation under diverse conditions.

\textit{2) Baselines and Metrics:} We compare our method with representative 3DGS-based SLAM systems, including MonoGS~\cite{matsuki2024gaussian_monoGs}, SplaTAM~\cite{keetha2024splatam}, GS-LIVM~\cite{xie2025gs}, and Gaussian-LIC2~\cite{lang2025gaussian2_gaussian_lic2}. For localization evaluation, we further include FAST-LIVO2~\cite{zheng2024fast}. As MonoGS and SplaTAM cannot directly operate on streaming sensor input, they are evaluated on recorded bag data in offline settings, while all other methods are evaluated under real-time input streaming.
Mapping quality is assessed using Peak Signal to Noise Ratio (PSNR), Structural Similarity (SSIM), and Learned Perceptual Image Patch Similarity (LPIPS), while localization accuracy is evaluated using Absolute Trajectory Error (ATE) in RMSE. For rendering evaluation, all reported results of our method are measured only on the held-out test images, which are strictly separated from the training images used for optimization. To directly quantify online efficiency, we define the Real-time factor as the total optimization runtime divided by the duration of the input sensor stream.

\textit{3) Implementation Details:} All experiments are conducted on the same platform for fair comparison, equipped with an Intel Core i9-14900KS CPU, an NVIDIA RTX 4090 GPU, and 128\,GB RAM, running Ubuntu 20.04 with ROS Noetic. The core modules are implemented in C++ and CUDA.

\subsection{Results Evaluation}

\input{Figures_tex/render_compare_1}
\input{Tables_tex/merge_table.tex}

\textit{1) Evaluation of Rendering:} Quantitative and qualitative evaluations are presented in Table~\ref{tab:rendering_runtime_results} and Fig.~\ref{fig:render_compare_all}, respectively. GS-LIVM over-relies on sweep reconstruction to enforce strict stream-synchronization. The assumption does not hold consistently in our sequences, resulting in degraded pose estimation and reconstruction quality. For evaluating SplaTAM and MonoGS, the LiDAR measurements have to be converted into depth images, which are inherently sparse and noisy after projection. Since SplaTAM exploits depth information for pose optimization and also relies on depth for Gaussian initialization, it is more sensitive to the quality of the input depth images, resulting in inferior reconstruction quality. In contrast, MonoGS mainly relies on photometric residuals for tracking and is therefore less directly affected by projected depth quality. Our experiments show that MonoGS still achieves competitive reconstruction quality among the compared methods. However, in highly unstructured road scenes, MonoGS incurs substantially higher runtime than the proposed method. Gaussian-LIC2 benefits from robust pose estimation and incremental Gaussian mapping, enabling a favorable trade-off between reconstruction quality and efficiency. However, its lack of loop closure limits global consistency in large-scale scenes. This is evident in the first row of Fig.~\ref{fig:render_compare_all}, where the reconstructed road markings overlap relative to the ground-truth image. In addition, the continuously expanding Gaussian map over long trajectories and incoming novel views increases the optimization load on the back end. Overall, the proposed method achieves the best or near-best performance on almost all evaluated sequences. 

As shown in Fig.~\ref{fig:render_compare_all}, our method renders sharper RGB images with fewer artifacts under novel views. The first row further indicates that the introduced Gaussian-based loop closure improves global Gaussian map consistency in revisited large-scale scenes, leading to a more consistent reconstruction than the compared methods. The last two rows show that, with the proposed cascaded initialization, our method provides better attribute initialization for subsequent optimization, thereby enabling more faithful recovery of view-dependent lighting and shading effects.

\textit{2) Evaluation of Localization:} Table~\ref{tab:pose_results} summarizes the quantitative results, showing that the proposed method consistently achieves the best overall performance on all sequences. By further introducing Gaussian-based loop closure, our method achieves better global consistency and reduced accumulated drift. As illustrated in Fig.~\ref{fig:traj_vis}, this advantage is particularly evident in the smaller drift along the vertical ($z$-axis) direction. Moreover, our method achieves better trajectory closure and geometric consistency after a full traversal and revisit of previously mapped areas, whereas most baseline methods still exhibit residual offsets or heading inconsistencies. These results indicate that the proposed method not only improves localization accuracy but also enhances the global consistency of the 3D Gaussian map.

\input{Tables_tex/ATE_only}
\input{Figures_tex/traj_vis}




\subsection{Runtime Analysis}

We benchmark the runtime performance of our proposed method from both system-level and module-level perspectives. Real-time operation is essential for long-term reliability and downstream applications such as online perception and autonomous navigation. An ideal system should continuously keep pace with incoming sensor streams for ego-pose estimation and 3D Gaussian mapping.

\textit{1) System-level runtime:} As shown in Table~\ref{tab:rendering_runtime_results}, the proposed method achieves Real-time factors close to $1$ on most sequences, indicating near-real-time full-pipeline processing while maintaining strong rendering quality. In contrast, several baseline methods yield substantially larger factors, and some even fail to complete the full sequence. This is due to their heavier optimization overhead for pose refinement and Gaussian map updating, which makes it difficult to continuously keep pace with the incoming sensor stream.

\input{Figures_tex/runtime_analysis}

\textit{2) Module-level runtime:} We consider two representative sequences with markedly different mapping scales: \textit{HKU Campus}, a small-scale sequence with a trajectory length of 64m, and \textit{Driving1}, a large-scale looped sequence with a trajectory length of 551m. Fig.~\ref{fig:runtime_analysis} illustrates both the temporal evolution and the overall runtime composition across these two scenes. In \textit{HKU Campus}, runtime is mainly dominated by state estimation and Gaussian primitives optimization. In \textit{Driving1}, however, loop closure consumes the majority of computation once triggered, accompanied by increased GPU memory usage. Thus, the dominant computational cost is scene dependent: in the small-scale case it is shared by front-end estimation and Gaussian optimization, whereas in the large-scale case it shifts to loop closure. Despite these differences, the system meets real-time requirements in both scenarios, achieving real-time factors of 1.03 and 1.01 on \textit{HKU Campus} and \textit{Driving1}, respectively. Overall, these results confirm stable real-time performance across different scene scales.

\section{ABLATION STUDY}

\input{Tables_tex/Ablation_Study}

We conduct an ablation study on the representative sequences \textit{HKU Campus} and \textit{Driving1} to evaluate the effectiveness of the proposed components under different scene scales. In the our framework, the feed-forward branch is instantiated with G$^3$Splat~\cite{g3splat}. As summarized in Table~\ref{tab:ablation_rendering}, the full system outperforms pure point-cloud initialization and the variants without the FFM or voxel-PCA, demonstrating the effectiveness of the cascaded initialization. In particular, removing the FFM leads to a larger drop in SSIM and LPIPS than in PSNR, suggesting that feed-forward predictions mainly improve structural and perceptual fidelity rather than average pixel-wise agreement. As our FFM is a modular component, we further compare two recent state-of-the-art feed-forward models, DepthSplat~\cite{xu2025depthsplat} and G$^3$Splat~\cite{g3splat}. While DepthSplat was able to reduce LPIPS but at the cost of PSNR and SSIM, G$^3$Splat was able to deliver better overall performance and is therefore adopted in the our framework. Moreover, the results show that loop closure significantly improves rendering performance in the large-scale looped scene, while introducing no noticeable degradation in the smaller sequence despite the extra computational overhead.

\section{CONCLUSIONS}
We present RMGS-SLAM, a tightly coupled LIV 3DGS-based SLAM system. By combining FFM with voxel-PCA priors, the proposed cascaded Gaussian initialization produces geometrically reliable primitives, which accelerates optimization convergence and supports real-time 3DGS mapping. Furthermore, the introduced 3DGS-based loop closure enhances both pose estimation and map consistency. Extensive experiments demonstrate that RMGS-SLAM achieves a strong balance among real-time performance, localization accuracy, and rendering quality compared with state-of-the-art methods. However, the current system still relies on high-performance hardware to sustain stable real-time operation in large-scale environments. Future work will focus on improving computational efficiency and extending its applicability to downstream robotic tasks such as navigation and grasping.

\bibliographystyle{IEEEtran}
\bibliography{ref}
\end{document}

%% file: Figures_tex/overview_figure.tex
\begin{figure*}[t]
    \centering
    \includegraphics[width=\textwidth]{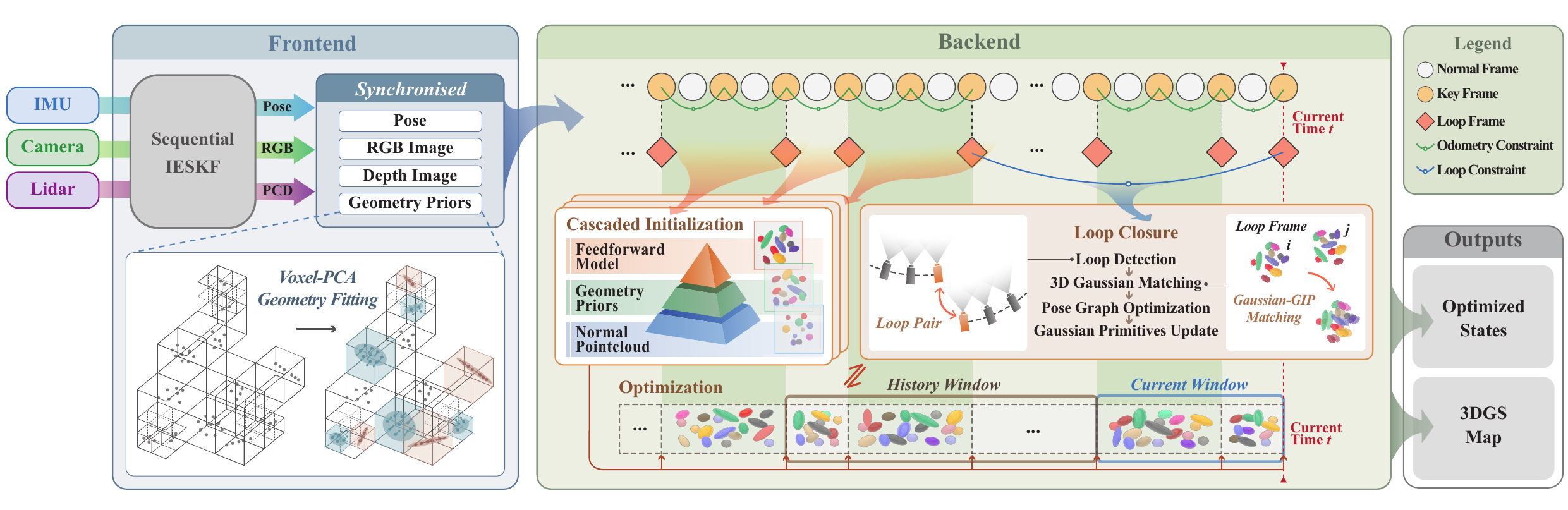}
    \caption{Overview of the proposed system. The framework follows a four-module design. (i) A LIV front-end developed upon a Sequential IESKF system~\cite{zheng2024fast} first estimates ego-motion and produces time-synchronized poses, RGB images, projected depth images, and voxel-PCA geometric priors. (ii) These outputs are then used to initialize 3D Gaussian primitives through a cascaded strategy that combines feed-forward predictions with geometry-aware priors. (iii) The global Gaussian map is then optimized asynchronously under photometric, structural, and geometric constraints. (iv) To improve global consistency, loop closure is performed directly on the Gaussian representation via Gaussian GICP, and the resulting loop constraints are jointly optimized with odometry in a pose graph.}
    \label{fig:system_overview}
\end{figure*}


%% file: Figures_tex/cascaded_init.tex
\begin{figure}[t]
    \centering
    \includegraphics[width=\columnwidth, height=6.0cm]{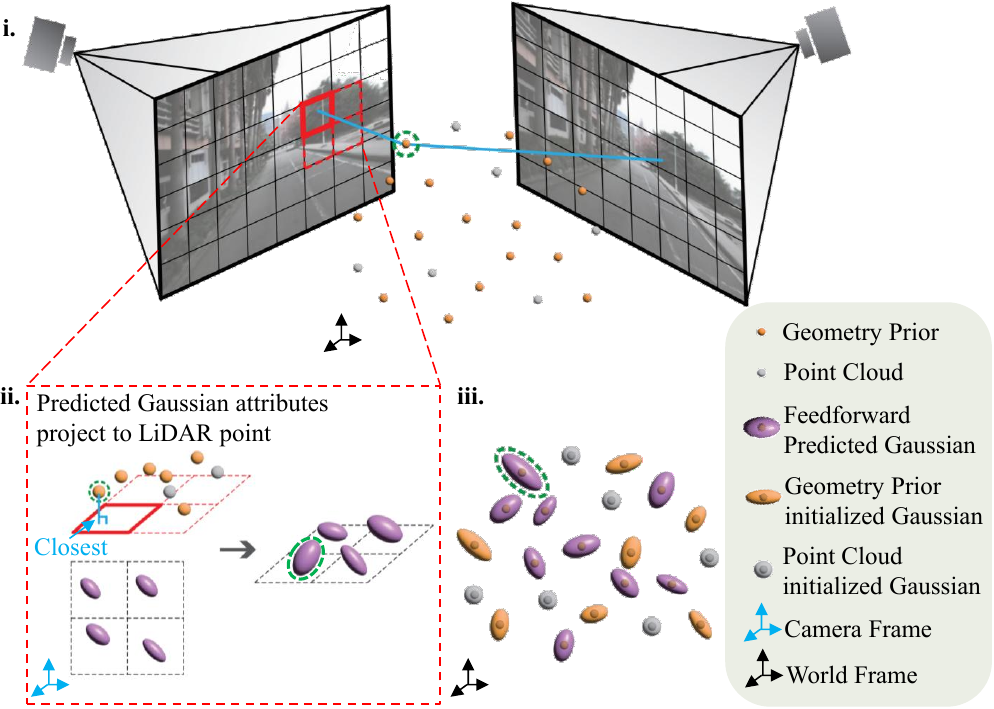}
    \caption{Cascaded Gaussian Primitive Initialization. (i) Each point is projected into the previous and current loopframe images, and the closer valid view is used. (ii) On the predicted per-pixel Gaussian attribute maps, only the closest point is retained at each pixel and its attributes are bilinearly interpolated. (iii) Model-initialized points are transformed to the world frame and the rest are initialized from voxel-PCA priors or an isotropic heuristic.}
    \label{fig:cascaded_init}
\end{figure}

%% file: Figures_tex/render_compare_1.tex
\begin{figure*}[t]
\centering
\setlength{\tabcolsep}{-0.1 pt} 
\renewcommand{\arraystretch}{1.0}
\providecommand{\imgw}{0.165\textwidth}
\providecommand{\qimg}[1]{\includegraphics[width=\imgw]{#1}}

\begin{tabular}{cccccc}
\textbf{GS-LIC2~\cite{lang2025gaussian2_gaussian_lic2}} & \textbf{GS-LIVM~\cite{xie2025gs}} & \textbf{MonoGS~\cite{matsuki2024gaussian_monoGs}} & \textbf{SplaTAM~\cite{keetha2024splatam}} & \textbf{Ours} & \textbf{Ground Truth} \\[2pt]

\qimg{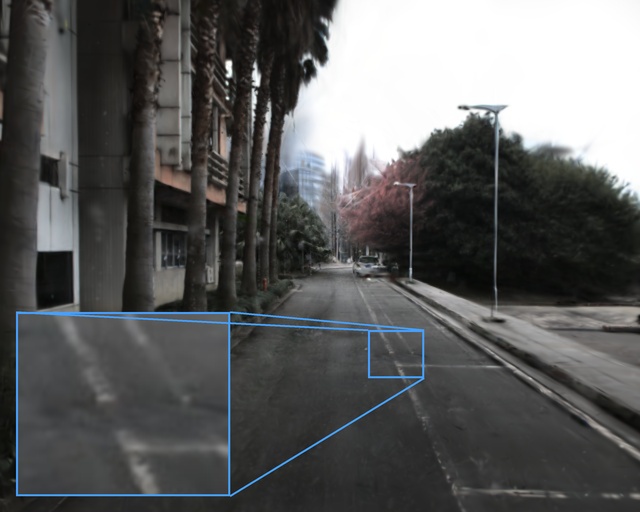} &
\qimg{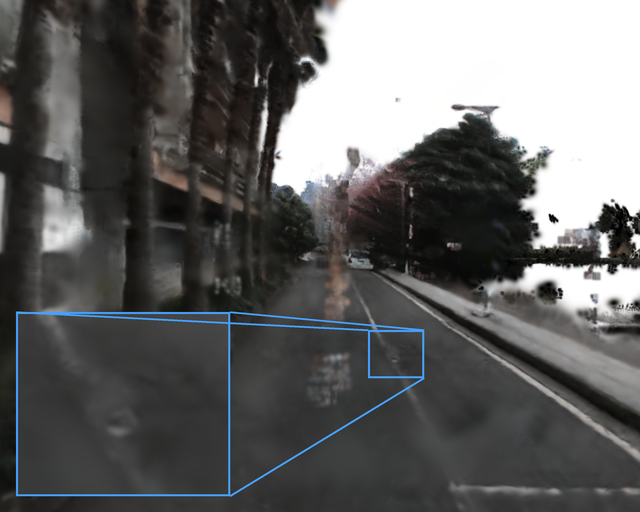} &
\qimg{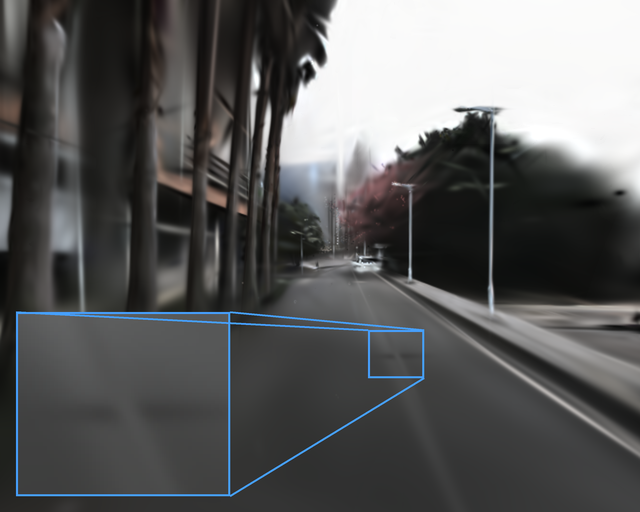} &
\qimg{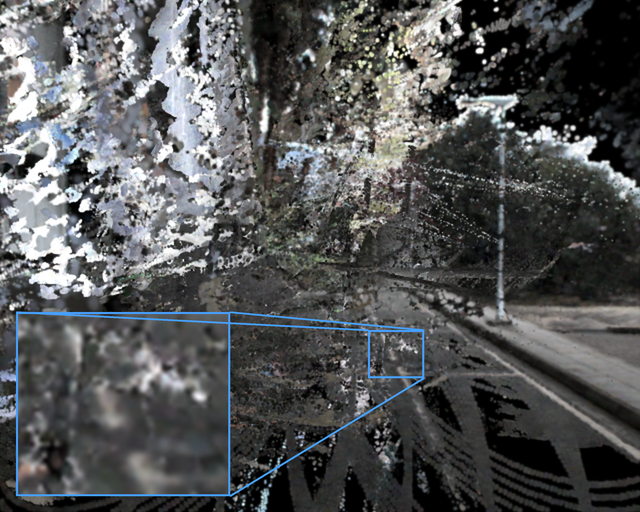} &
\qimg{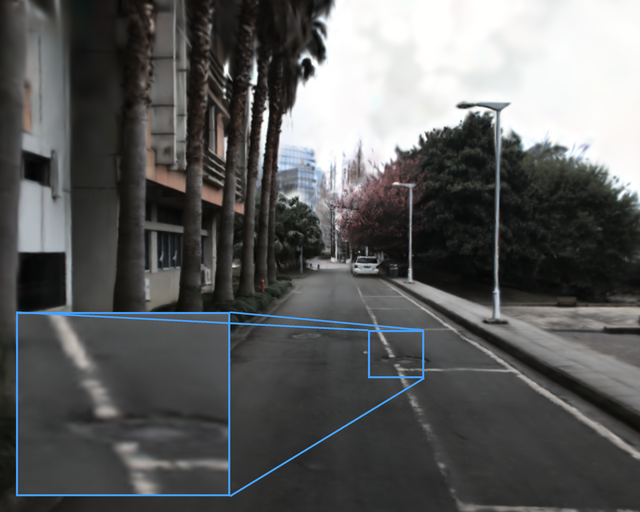} &
\qimg{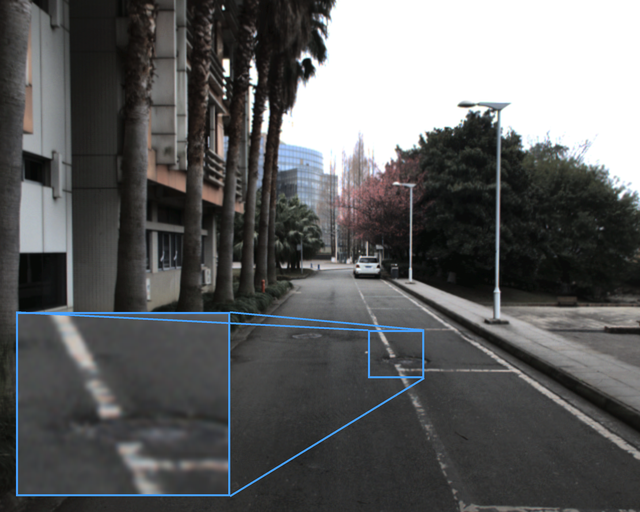} \\[-3.5pt]

\qimg{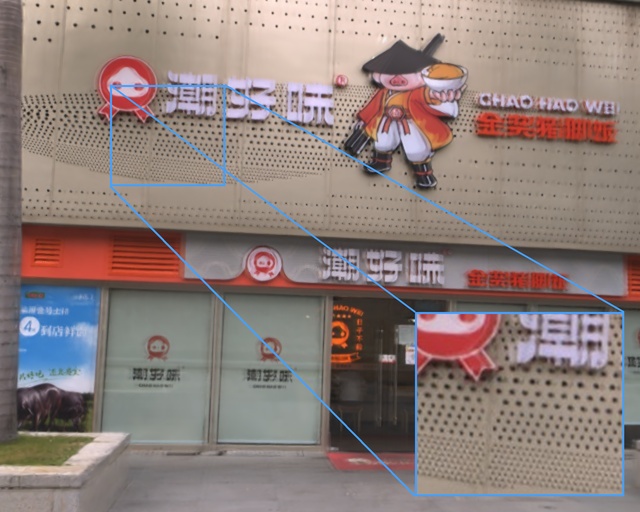} &
\qimg{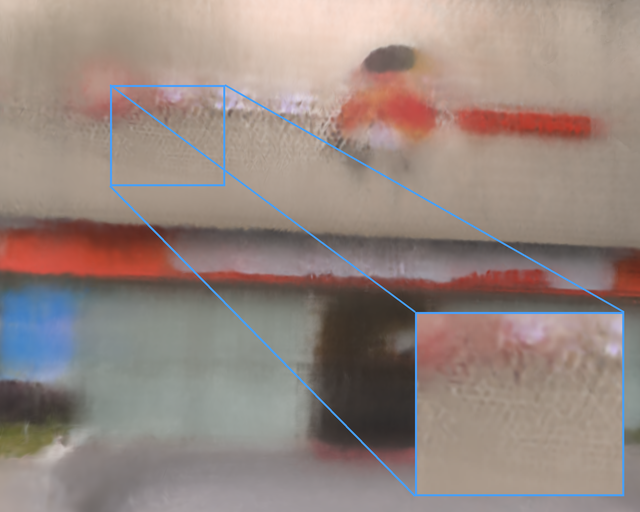} &
\qimg{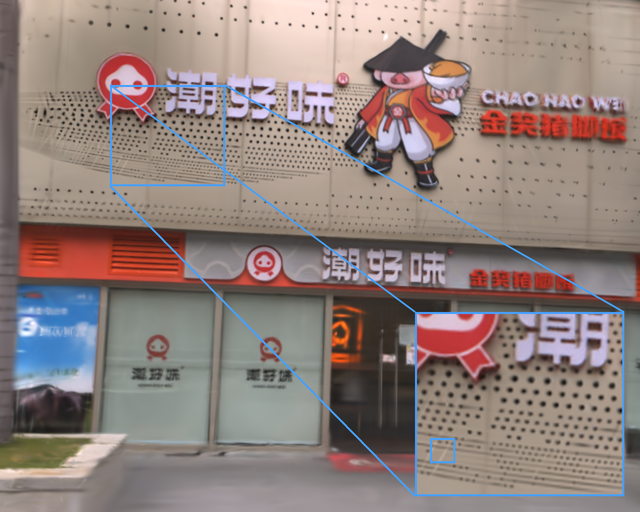} &
\qimg{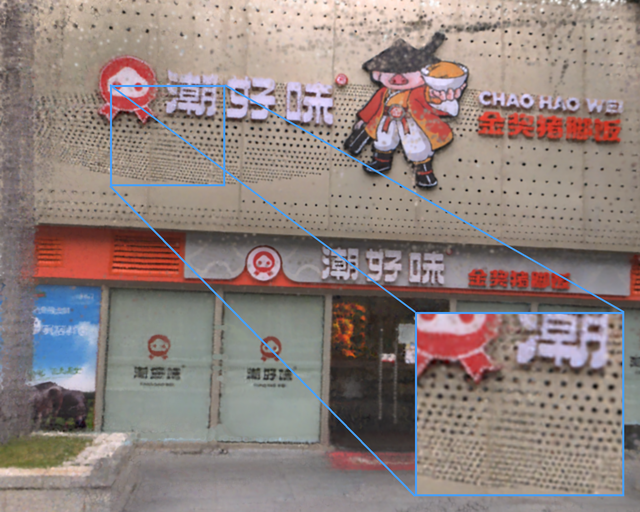} &
\qimg{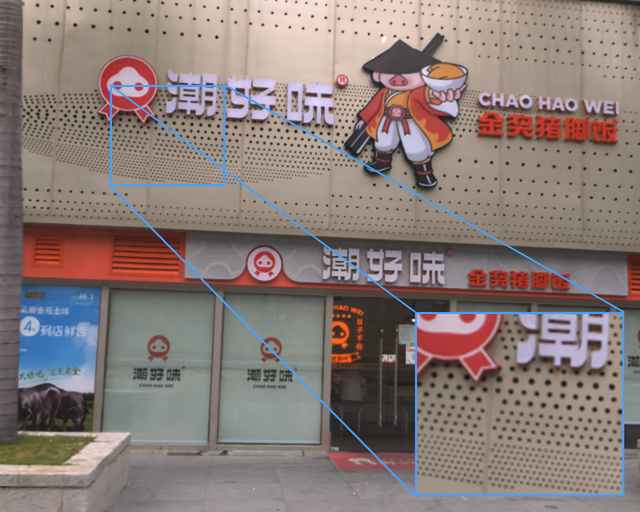} &
\qimg{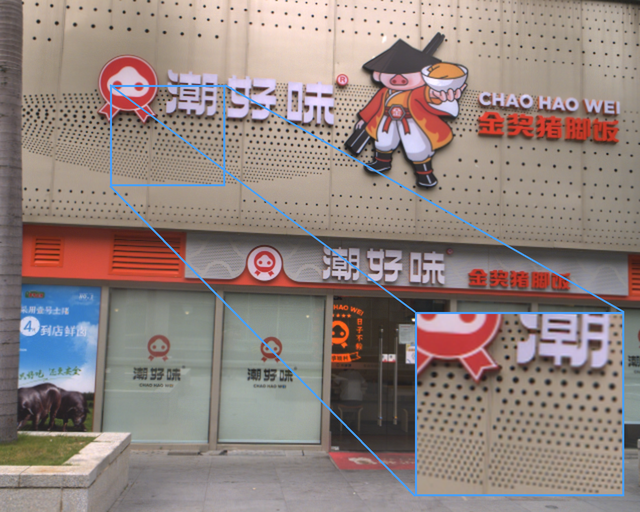} \\[-3.5pt]

\qimg{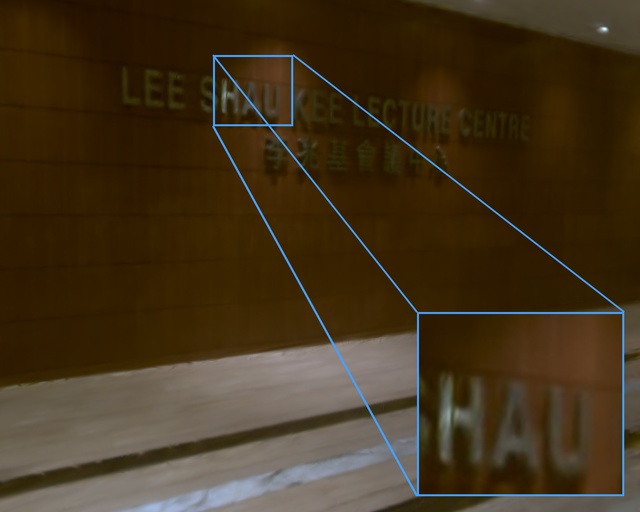} &
\qimg{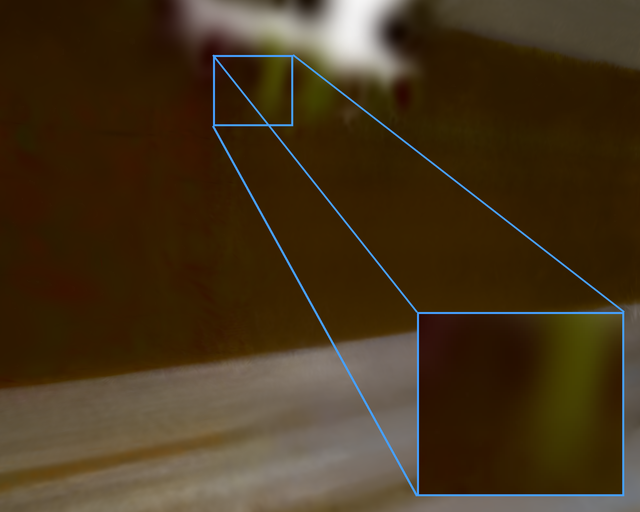} &
\qimg{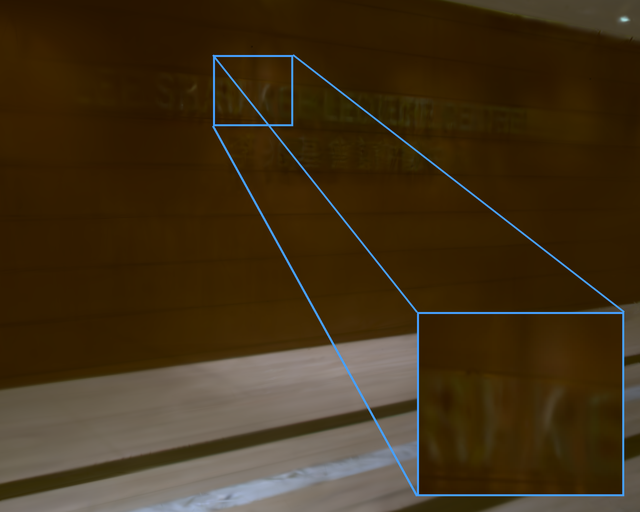} &
\qimg{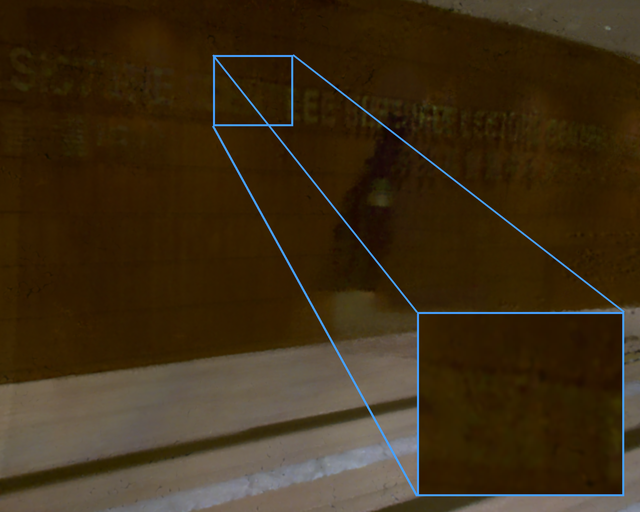} &
\qimg{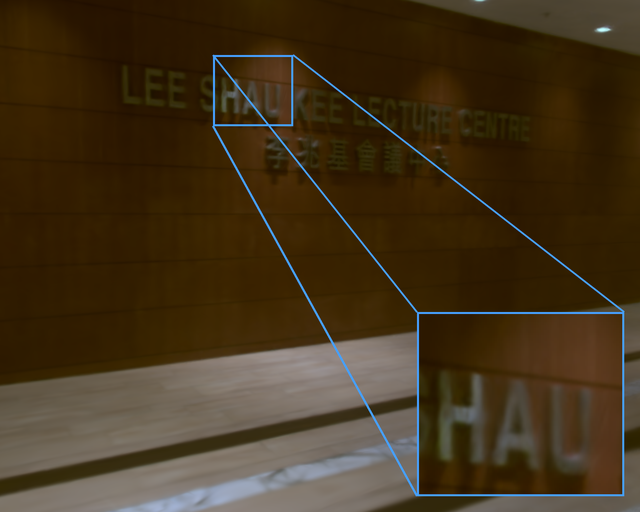} &
\qimg{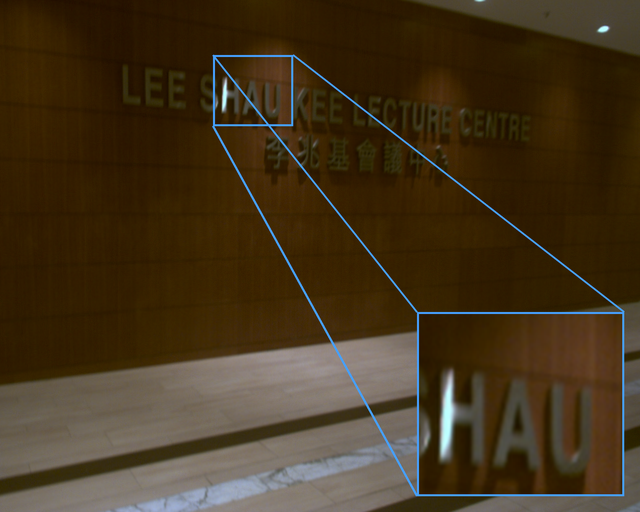} \\[-3.8pt]

\qimg{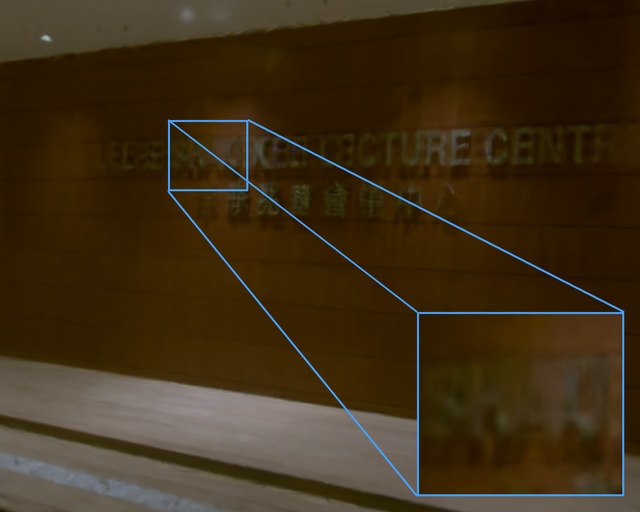} &
\qimg{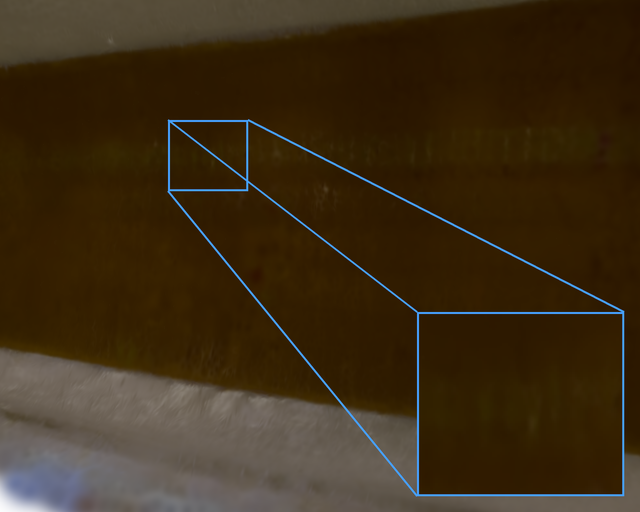} &
\qimg{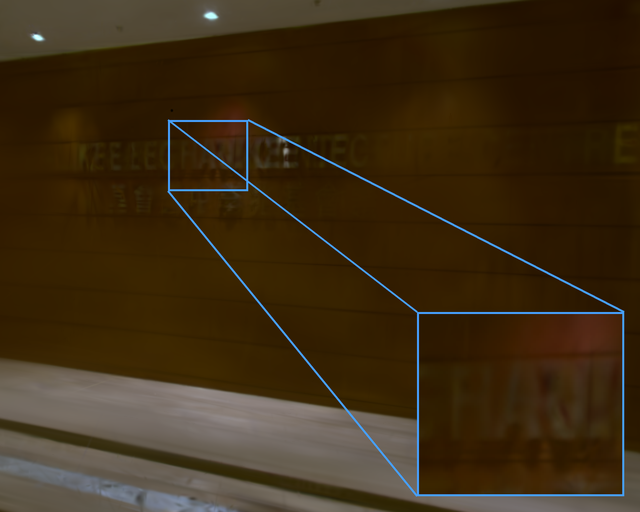} &
\qimg{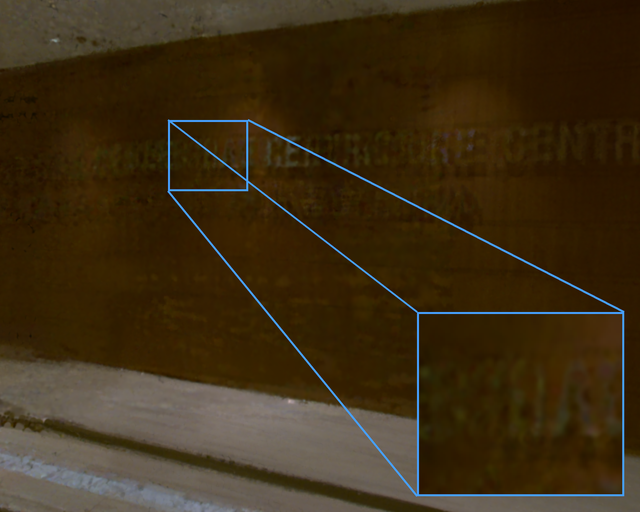} &
\qimg{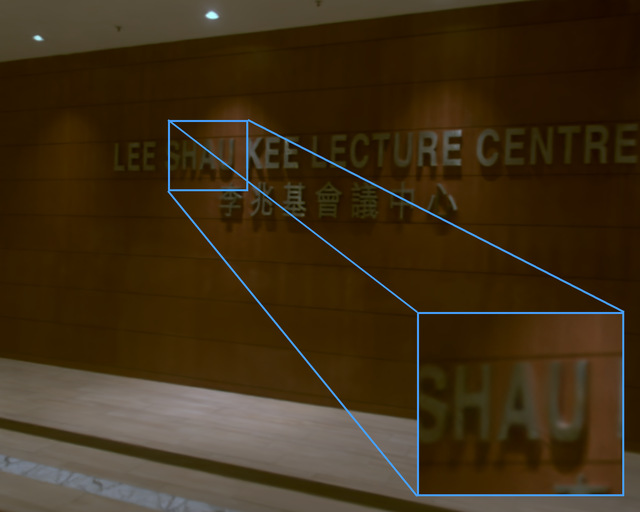} &
\qimg{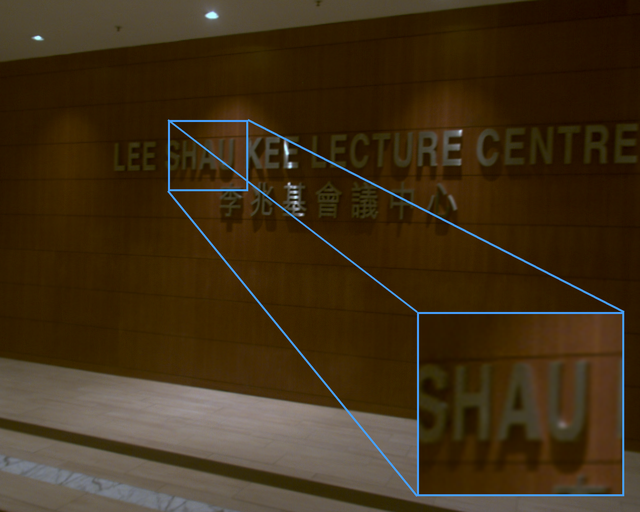} \\

\end{tabular}

\caption{Qualitative comparison of rendering results. The blue boxed region is enlarged and displayed in the corner for detailed visual inspection.}
\label{fig:render_compare_all}
\end{figure*}

%% file: Tables_tex/merge_table.tex
\begin{table*}[!t]
\centering
\caption{Rendering quality and runtime efficiency comparison on selected datasets.}
\label{tab:rendering_runtime_results}
\footnotesize
\renewcommand{\arraystretch}{1.08}
\setlength{\tabcolsep}{4.0pt}
\begin{tabular}{ll*{8}{C{1.30cm}}}
\toprule
Method & Metric
& \shortstack{Lecture\\Center01}
& \shortstack{HKU\\Campus}
& \shortstack{Retail\\Street}
& \shortstack{CBD\\Building02}
& \shortstack{Driving1}
& \shortstack{Driving2}
& \shortstack{HKairport\\03}
& \shortstack{HKisland\\03} \\
\midrule

\multirow{4}{*}{Gaussian-LIC2~\cite{lang2025gaussian2_gaussian_lic2}}
& PSNR$\uparrow$
& 31.21 
& 27.06 
& 24.64 
& 23.16 
& 22.09 
& \textbf{23.72} 
& 14.84 
& 17.65 \\ 

& SSIM$\uparrow$
& 0.91 
& 0.76 
& 0.77 
& 0.77 
& 0.76 
& 0.81 
& 0.36 
& 0.44 
\\ 

& LPIPS$\downarrow$
& 0.17 
& 0.21 
& 0.76 
& 0.23 
& 0.28 
& \textbf{0.22} 
& 0.59 
& 0.54 
\\ 

& Real-time factor$\downarrow$
& 1.14 
& 1.09 
& 1.05 
& 1.17 
& 1.08 
& 2.14 
& $\infty$ 
& $\infty$ \\
\cdashline{1-10}

\multirow{4}{*}{GS-LIVM~\cite{xie2025gs}}
& PSNR$\uparrow$
& 21.48 
& 12.06 
& 17.34 
& 13.57 
& 16.17 
& 14.86 
& $\times$
& 2.96 \\ 

& SSIM$\uparrow$
& 0.75 
& 0.42  
& 0.51  
& 0.51 
& 0.55 
& 0.56 
& $\times$
& 0.16 \\ 

& LPIPS$\downarrow$
& 0.56 
& 0.76 
& 0.70 
& 0.75 
& 0.53 
& 0.53 
& $\times$
& 1.01 \\

& Real-time factor$\downarrow$
& \textbf{1.04} 
& \textbf{1.03} 
& \textbf{1.02} 
& \textbf{1.01} 
& 1.02 
& \textbf{1.01} 
& $\infty$
& \textbf{1.01}\\ 
\cdashline{1-10}

\multirow{4}{*}{MonoGS~\cite{matsuki2024gaussian_monoGs}}
& PSNR$\uparrow$
& 30.22 
& 30.12 
& 25.53 
& 23.04 
& \textbf{23.79} 
& 22.17 
& 16.55 
& 13.87 \\ 

& SSIM$\uparrow$
& 0.91 
& 0.84 
& 0.81 
& 0.74 
& 0.76 
& 0.72 
& 0.38 
& 0.37 \\ 

& LPIPS$\downarrow$
& 0.24 
& 0.22 
& 0.17 
& 0.33 
& 0.37 
& 0.45 
& 0.56 
& 0.65 \\ 

& Real-time factor$\downarrow$
& 6.21 
& 5.98 
& 6.37 
& 5.90 
& 8.15 
& 8.24 
& 7.40 
& 3.06 \\ 
\cdashline{1-10}

\multirow{4}{*}{SplaTAM~\cite{keetha2024splatam}}
& PSNR$\uparrow$
& 25.23
& 20.57
& 20.84
& 11.89 
& 9.93
& 9.64
& 13.16
& 11.96 \\ 

& SSIM$\uparrow$
& 0.81
& 0.65
& 0.66
& 0.32 
& 0.31
& 0.29
& 0.29
& 0.23 \\ 

& LPIPS$\downarrow$
& 0.33
& 0.33
& 0.26
& 0.70 
& 0.65
& 0.67
& 0.62
& 0.74 \\ 

& Real-time factor$\downarrow$
& 38.30
& 22.41
& 15.86
& 8.62 
& 2.71
& 4.29 
& 6.30 
& 5.46 \\ 
\cdashline{1-10}

\multirow{4}{*}{RMGS-SLAM (ours)}
& PSNR$\uparrow$
& \textbf{35.63}
& \textbf{30.85}
& \textbf{28.44}
& \textbf{27.64}
& 21.79
& 22.20
& \textbf{19.64} 
& \textbf{19.55} \\ 

& SSIM$\uparrow$
& \textbf{0.95}
& \textbf{0.88}
& \textbf{0.91}
& \textbf{0.93}
& \textbf{0.80} 
& \textbf{0.82}
& \textbf{0.54} 
& \textbf{0.67} \\ 

& LPIPS$\downarrow$
& \textbf{0.11}
& \textbf{0.12}
& \textbf{0.07}
& \textbf{0.09}
& \textbf{0.24} 
& 0.23
& \textbf{0.33} 
& \textbf{0.37} \\ 

& Real-time factor$\downarrow$
& \textbf{1.04}
& \textbf{1.03}
& \textbf{1.02} 
& \textbf{1.01}
& \textbf{1.01}
& \textbf{1.01}
& \textbf{1.50} 
& \textbf{1.01} \\
\bottomrule
\end{tabular}

\vspace{0.8mm}
\begin{minipage}{\textwidth}
\raggedright
\scriptsize
\textit{Note:} $\times$ indicates invalid renderings, and a real-time factor of $\infty$ indicates failure to complete the full sequence on our setup.
\end{minipage}
\end{table*}

%% file: Tables_tex/ATE_only.tex
\begin{table}[t]
\vspace{0.3cm}
\centering
\caption{Pose accuracy comparison on different datasets.}
\label{tab:pose_results}
\footnotesize
\renewcommand{\arraystretch}{0.98}
\setlength{\tabcolsep}{4pt}
\begin{tabular}{lcccc}
\toprule
Method & Driving1 & Driving2 & HKairport03 & HKisland03 \\
\midrule
FAST-LIVO2~\cite{zheng2024fast}
& 3.55 
& 2.86 
& 1.41 
& 2.74\\ 
Gaussian-LIC2~\cite{lang2025gaussian2_gaussian_lic2}
& 3.66 
& 2.81 
& $\times$ 
& $\times$ \\
GS-LIVM~\cite{xie2025gs}
& 4.86 
& 2.95 
& $\times$ 
& 65.26 
\\ 
MonoGS~\cite{matsuki2024gaussian_monoGs}
& 6.75 
& 13.08 
& 106.42 
& 140.87 \\ 
SplaTAM~\cite{keetha2024splatam}
& 63.32 
& 108.12 
& 117.12 
& 156.63 
\\ 
\midrule
RMGS-SLAM (ours)
& \textbf{0.41} 
& \textbf{1.55} 
& \textbf{1.39} 
& \textbf{2.72} \\ 
\bottomrule
\end{tabular}

\vspace{0.6mm}
\begin{minipage}{\columnwidth}
\raggedright
\scriptsize
\textit{Note:} $\times$ denotes that the method failed to complete the full sequence.
\end{minipage}
\end{table}

%% file: Figures_tex/traj_vis.tex
\begin{figure}[!htbp]
    \centering
    \includegraphics[width=\columnwidth, height=7cm]{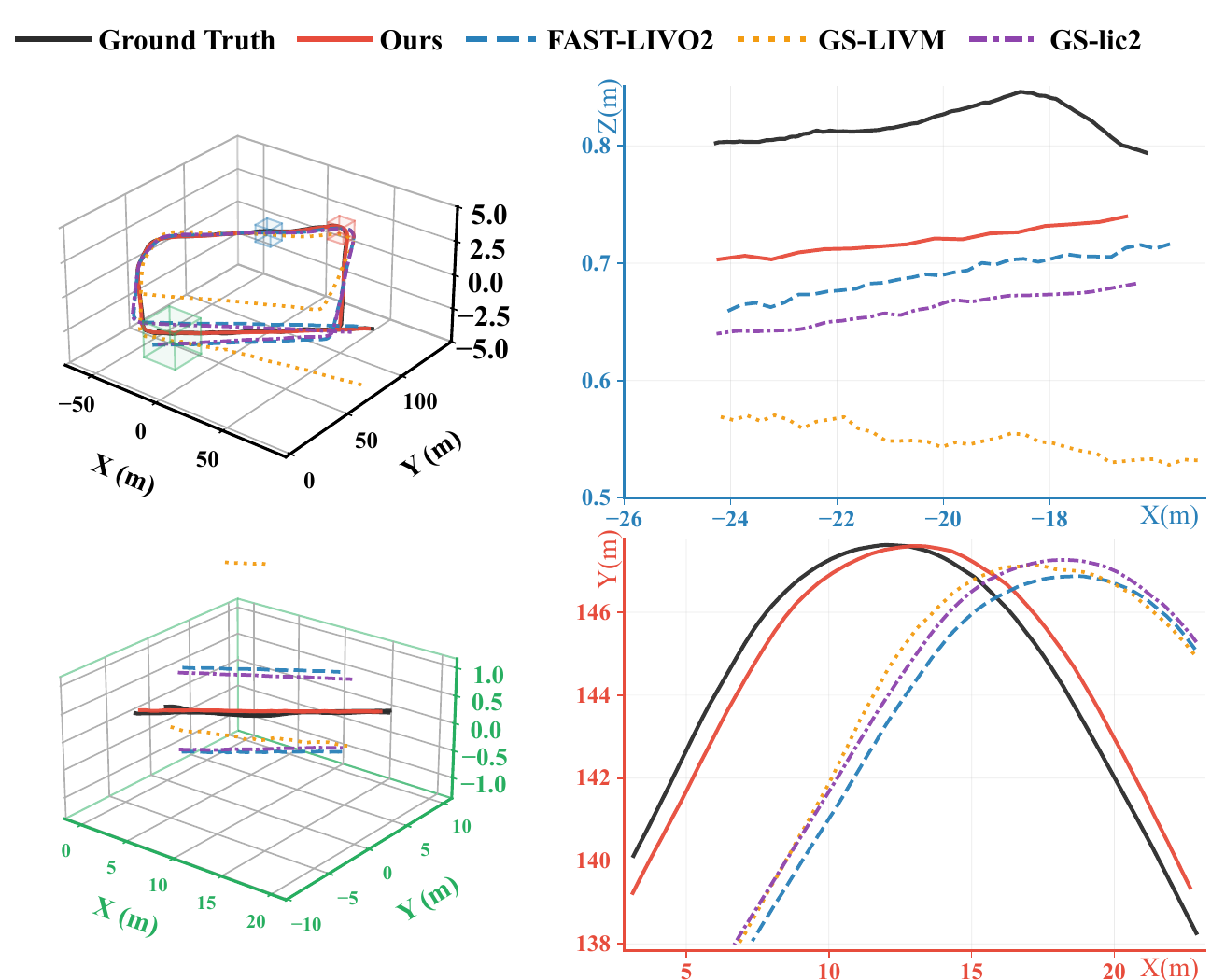}
    \caption{Trajectory comparison for the \textit{Driving1} sequence. The upper-left subfigure shows the overall 3D trajectory, and the others show local zoomed-in views at approximately $8.3\times$ magnification.}
    \label{fig:traj_vis}
\end{figure}

%% file: Figures_tex/runtime_analysis.tex

\begin{figure}[t]
    \centering
    \includegraphics[width=\columnwidth, height=6cm]{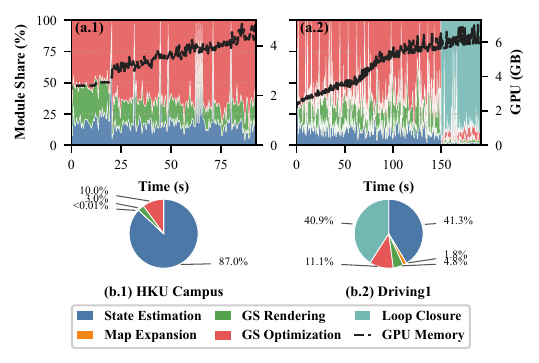}
    \caption{Runtime analysis across small (1, HKU Campus) and large (2, Driving1) squences. The subfigure (a) presents compute time and memory consumption of each module sampled at 0.5\,s intervals. The subfigure (b) summarizes the corresponding time-weighted overall runtime percentage of each module.}
    \label{fig:runtime_analysis}
\end{figure}


%% file: Tables_tex/Ablation_Study.tex
\begin{table}[t]
\centering
\caption{Ablation study of different component choices.}
\label{tab:ablation_rendering}
\footnotesize
\renewcommand{\arraystretch}{0.98}
\setlength{\tabcolsep}{2pt}
\begin{tabular}{lcccccc}
\toprule
\multirow{2}{*}{\textbf{Variant}}
& \multicolumn{3}{c}{\textbf{HKU Campus(small)}}
& \multicolumn{3}{c}{\textbf{Driving1(large)}} \\
\cmidrule(l{0.25em}r{0.25em}){2-4}
\cmidrule(l{0.25em}r{0.25em}){5-7}
& PSNR$\uparrow$ & SSIM$\uparrow$ & LPIPS$\downarrow$
& PSNR$\uparrow$ & SSIM$\uparrow$ & LPIPS$\downarrow$ \\
\midrule
w/o voxel-PCA \& FFM
& 27.53 & 0.82 & 0.15
& 19.57 & 0.73 & 0.30 \\
w/o FFM
& 29.77 & 0.85 & 0.17
& \textbf{21.81} & 0.76 & 0.29 \\
w/o voxel-PCA
& 27.98 & 0.81 & 0.16
& 20.76 & 0.75 & 0.27 \\
\noalign{\vskip 1.5pt}
\cdashline{1-7}[2pt/2pt]
\noalign{\vskip 2pt}

w/ FFM (DepthSplat)
& 28.91 
& 0.83 
& 0.15
& 20.77 
& 0.75 
& 0.26 \\
\noalign{\vskip 1.5pt}
\cdashline{1-7}[2pt/2pt]
\noalign{\vskip 2pt}

w/o loop closure
& \textbf{30.88} & \textbf{0.88} & \textbf{0.11}
& 19.82 & 0.74 & 0.30 \\
\midrule
RMGS-SLAM (ours)
& 30.85 & \textbf{0.88} & 0.12
& 21.79 & \textbf{0.80} & \textbf{0.24} \\
\bottomrule
\end{tabular}
\end{table}

%% file: main.bbl
\begin{thebibliography}{10}
\providecommand{\url}[1]{#1}
\csname url@samestyle\endcsname
\providecommand{\newblock}{\relax}
\providecommand{\bibinfo}[2]{#2}
\providecommand{\BIBentrySTDinterwordspacing}{\spaceskip=0pt\relax}
\providecommand{\BIBentryALTinterwordstretchfactor}{4}
\providecommand{\BIBentryALTinterwordspacing}{\spaceskip=\fontdimen2\font plus
\BIBentryALTinterwordstretchfactor\fontdimen3\font minus \fontdimen4\font\relax}
\providecommand{\BIBforeignlanguage}[2]{{%
\expandafter\ifx\csname l@#1\endcsname\relax
\typeout{** WARNING: IEEEtran.bst: No hyphenation pattern has been}%
\typeout{** loaded for the language `#1'. Using the pattern for}%
\typeout{** the default language instead.}%
\else
\language=\csname l@#1\endcsname
\fi
#2}}
\providecommand{\BIBdecl}{\relax}
\BIBdecl

\bibitem{cadena2017past}
C.~Cadena, L.~Carlone, H.~Carrillo, Y.~Latif, D.~Scaramuzza, J.~Neira, I.~Reid, and J.~J. Leonard, ``Past, present, and future of simultaneous localization and mapping: Toward the robust-perception age,'' \emph{IEEE Transactions on robotics}, vol.~32, no.~6, pp. 1309--1332, 2017.

\bibitem{zhang2014loam}
J.~Zhang, S.~Singh \emph{et~al.}, ``Loam: Lidar odometry and mapping in real-time.'' in \emph{Robotics: Science and systems}, vol.~2, no.~9.\hskip 1em plus 0.5em minus 0.4em\relax Berkeley, CA, 2014, pp. 1--9.

\bibitem{mur2017orb}
R.~Mur-Artal and J.~D. Tard{\'o}s, ``Orb-slam2: An open-source slam system for monocular, stereo, and rgb-d cameras,'' \emph{IEEE transactions on robotics}, vol.~33, no.~5, pp. 1255--1262, 2017.

\bibitem{zheng2024fast}
C.~Zheng, W.~Xu, Z.~Zou, T.~Hua, C.~Yuan, D.~He, B.~Zhou, Z.~Liu, J.~Lin, F.~Zhu \emph{et~al.}, ``Fast-livo2: Fast, direct lidar--inertial--visual odometry,'' \emph{IEEE Transactions on Robotics}, vol.~41, pp. 326--346, 2024.

\bibitem{kerbl3Dgaussians}
B.~Kerbl, G.~Kopanas, T.~Leimk{\"u}hler, and G.~Drettakis, ``3d gaussian splatting for real-time radiance field rendering,'' \emph{ACM Transactions on Graphics}, vol.~42, no.~4, July 2023.

\bibitem{zhou2024drivinggaussian}
X.~Zhou, Z.~Lin, X.~Shan, Y.~Wang, D.~Sun, and M.-H. Yang, ``Drivinggaussian: Composite gaussian splatting for surrounding dynamic autonomous driving scenes,'' in \emph{Proceedings of the IEEE/CVF conference on computer vision and pattern recognition}, 2024, pp. 21\,634--21\,643.

\bibitem{huang20242d}
B.~Huang, Z.~Yu, A.~Chen, A.~Geiger, and S.~Gao, ``2d gaussian splatting for geometrically accurate radiance fields,'' in \emph{ACM SIGGRAPH 2024 conference papers}, 2024, pp. 1--11.

\bibitem{yan2023gs_gsslam}
C.~Yan, D.~Qu, D.~Xu, B.~Zhao, Z.~Wang, D.~Wang, and X.~Li, ``Gs-slam: Dense visual slam with 3d gaussian splatting,'' in \emph{CVPR}, 2024.

\bibitem{matsuki2024gaussian_monoGs}
H.~Matsuki, R.~Murai, P.~H. Kelly, and A.~J. Davison, ``Gaussian splatting slam,'' in \emph{Proceedings of the IEEE/CVF conference on computer vision and pattern recognition}, 2024, pp. 18\,039--18\,048.

\bibitem{keetha2024splatam}
N.~Keetha, J.~Karhade, K.~M. Jatavallabhula, G.~Yang, S.~Scherer, D.~Ramanan, and J.~Luiten, ``Splatam: Splat track \& map 3d gaussians for dense rgb-d slam,'' in \emph{Proceedings of the IEEE/CVF conference on computer vision and pattern recognition}, 2024, pp. 21\,357--21\,366.

\bibitem{deng2025vpgs}
T.~Deng, W.~Wu, J.~He, Y.~Pan, X.~Jiang, S.~Yuan, D.~Wang, H.~Wang, and W.~Chen, ``Vpgs-slam: Voxel-based progressive 3d gaussian slam in large-scale scenes,'' \emph{arXiv preprint arXiv:2505.18992}, 2025.

\bibitem{hong2025gs}
S.~Hong, C.~Zheng, Y.~Shen, C.~Li, F.~Zhang, T.~Qin, and S.~Shen, ``Gs-livo: Real-time lidar, inertial, and visual multi-sensor fused odometry with gaussian mapping,'' \emph{IEEE Transactions on Robotics}, 2025.

\bibitem{phan2025fusiongs}
T.-D. Phan and G.-W. Kim, ``Fusiongs-slam: Multiple sensors fusion for localization and real-time photorealistic mapping,'' \emph{IEEE Robotics and Automation Letters}, 2025.

\bibitem{lang2025gaussian_gaussian_lic}
X.~Lang, L.~Li, C.~Wu, C.~Zhao, L.~Liu, Y.~Liu, J.~Lv, and X.~Zuo, ``Gaussian-lic: Real-time photo-realistic slam with gaussian splatting and lidar-inertial-camera fusion,'' in \emph{2025 IEEE International Conference on Robotics and Automation (ICRA)}.\hskip 1em plus 0.5em minus 0.4em\relax IEEE, 2025, pp. 8500--8507.

\bibitem{lang2025gaussian2_gaussian_lic2}
X.~Lang, J.~Lv, K.~Tang, L.~Li, J.~Huang, L.~Liu, Y.~Liu, and X.~Zuo, ``Gaussian-lic2: Lidar-inertial-camera gaussian splatting slam,'' \emph{arXiv}, 2025.

\bibitem{xie2025gs}
Y.~Xie, Z.~Huang, J.~Wu, and J.~Ma, ``Gs-livm: Real-time photo-realistic lidar-inertial-visual mapping with gaussian splatting,'' in \emph{Proceedings of the IEEE/CVF International Conference on Computer Vision}, 2025, pp. 26\,869--26\,878.

\bibitem{xu2025depthsplat}
H.~Xu, S.~Peng, F.~Wang, H.~Blum, D.~Barath, A.~Geiger, and M.~Pollefeys, ``Depthsplat: Connecting gaussian splatting and depth,'' in \emph{Proceedings of the Computer Vision and Pattern Recognition Conference}, 2025, pp. 16\,453--16\,463.

\bibitem{ye2024no}
B.~Ye, S.~Liu, H.~Xu, X.~Li, M.~Pollefeys, M.-H. Yang, and S.~Peng, ``No pose, no problem: Surprisingly simple 3d gaussian splats from sparse unposed images,'' \emph{arXiv preprint arXiv:2410.24207}, 2024.

\bibitem{g3splat}
M.~Hosseinzadeh \emph{et~al.}, ``Geometrically consistent generalizable gaussian splatting,'' \emph{arXiv preprint arXiv:2512.17547}, 2025.

\bibitem{cheng2025tls}
S.~Cheng, S.~He, F.~Duan, and N.~An, ``Tls-slam: Gaussian splatting slam tailored for large-scale scenes,'' \emph{IEEE Robotics and Automation Letters}, vol.~10, no.~3, pp. 2814--2821, 2025.

\bibitem{hong2024liv}
S.~Hong, J.~He, X.~Zheng, and C.~Zheng, ``Liv-gaussmap: Lidar-inertial-visual fusion for real-time 3d radiance field map rendering,'' \emph{IEEE Robotics and Automation Letters}, vol.~9, no.~11, pp. 9765--9772, 2024.

\bibitem{xiao2024liv}
R.~Xiao, W.~Liu, Y.~Chen, and L.~Hu, ``Liv-gs: Lidar-vision integration for 3d gaussian splatting slam in outdoor environments,'' \emph{IEEE Robotics and Automation Letters}, vol.~10, no.~1, pp. 421--428, 2024.

\bibitem{zhao2025lvi}
H.~Zhao, W.~Guan, and P.~Lu, ``Lvi-gs: Tightly-coupled lidar-visual-inertial slam using 3d gaussian splatting,'' \emph{IEEE Transactions on Instrumentation and Measurement}, 2025.

\bibitem{yuan2022efficient}
C.~Yuan, W.~Xu, X.~Liu, X.~Hong, and F.~Zhang, ``Efficient and probabilistic adaptive voxel mapping for accurate online lidar odometry,'' \emph{IEEE Robotics and Automation Letters}, vol.~7, no.~3, pp. 8518--8525, 2022.

\bibitem{liu2021balm}
Z.~Liu and F.~Zhang, ``Balm: Bundle adjustment for lidar mapping,'' \emph{IEEE Robotics and Automation Letters}, vol.~6, no.~2, pp. 3184--3191, 2021.

\bibitem{segal2009generalized}
A.~Segal, D.~Haehnel, S.~Thrun \emph{et~al.}, ``Generalized-icp.'' in \emph{Robotics: science and systems}, vol.~2, no.~4.\hskip 1em plus 0.5em minus 0.4em\relax Seattle, WA, 2009, p. 435.

\bibitem{kaess2012isam2}
M.~Kaess, H.~Johannsson, R.~Roberts, V.~Ila, J.~J. Leonard, and F.~Dellaert, ``isam2: Incremental smoothing and mapping using the bayes tree,'' \emph{International Journal of Robotics Research}, vol.~31, no.~2, pp. 216--235, 2012.

\bibitem{li2024mars}
H.~Li, Y.~Zou, N.~Chen, J.~Lin, X.~Liu, W.~Xu, C.~Zheng, R.~Li, D.~He, F.~Kong \emph{et~al.}, ``Mars-lvig dataset: A multi-sensor aerial robots slam dataset for lidar-visual-inertial-gnss fusion,'' \emph{The International Journal of Robotics Research}, vol.~43, no.~8, pp. 1114--1127, 2024.

\end{thebibliography}
